\documentclass[a4paper,fleqn]{cas-dc}

\usepackage[authoryear,longnamesfirst]{natbib}

\usepackage{booktabs}
\usepackage{algorithm}
\usepackage{algorithmic}
\usepackage{multirow}
\usepackage{array}
\usepackage{adjustbox}
\usepackage{pgfplots}
\pgfplotsset{compat=1.18}

\usepackage{xcolor}
\definecolor{cValidated}{HTML}{2E86AB}   
\definecolor{cFiltered}{HTML}{C4C4C4}    
\definecolor{cZeroLine}{HTML}{D64045}    

\usepackage{tikz}
\usetikzlibrary{arrows.meta,positioning,shapes.geometric,calc,fit,backgrounds}
\pgfdeclarelayer{bg}
\pgfsetlayers{bg,main}

\tikzset{
  box/.style={rectangle, rounded corners=2pt, draw, thick, align=center, minimum width=22mm, minimum height=7mm},
  bigbox/.style={rectangle, rounded corners=2pt, draw, thick, align=center, minimum width=32mm, minimum height=10mm},
  decision/.style={diamond, aspect=2, draw, thick, align=center, inner sep=1.2pt},
  group/.style={rectangle, rounded corners=3pt, draw=black!40, very thick, inner sep=6pt},
  arrow/.style={-{Latex[length=3mm]}, thick}
}

\usepackage{pifont}
\newcommand{\cmark}{\textcolor{green!70!black}{\ding{51}}}
\newcommand{\xmark}{\textcolor{red!70!black}{\ding{55}}}

\begin{document}
\let\WriteBookmarks\relax
\def\floatpagepagefraction{1}
\def\textpagefraction{.001}

\shorttitle{Refutation-Validated ABSA for Energy Market Returns}

\shortauthors{W. van der Heever et~al.}

\title [mode = title]{Beyond Correlation: Refutation-Validated Aspect-Based Sentiment Analysis for Explainable Energy Market Returns}



\author[1]{Wihan {van der Heever}}[orcid=0000-0003-2616-4313] 
\cormark[1]
\ead{wihan001@e.ntu.edu.sg}

\credit{Conceptualization, Methodology, Software, Formal Analysis, Writing - Original Draft, Visualization}

\author[3]{Keane Ong}[orcid=0009-0000-2397-1553]
\ead{keane.ongweiyang@u.nus.edu}

\credit{Data curation, Investigation}

\author[2]{Ranjan Satapathy}[orcid=0000-0002-0733-7381]
\ead{satapathy_ranjan@ihpc.a-star.edu.sg}

\credit{Methodology, Validation, Writing - Review \& Editing}

\author[1]{Erik Cambria}[orcid=0000-0002-3030-1280]
\cormark[2]
\ead{cambria@ntu.edu.sg}

\credit{Supervision, Resources, Writing - Review \& Editing}

\affiliation[1]{organization={College of Computing and Data Science, Nanyang Technological University},
            addressline={50 Nanyang Ave}, 
            city={Singapore},
            postcode={639798}, 
            country={Singapore}}

\affiliation[2]{organization={Institute of High Performance Computing, Agency for Science, Technology and Research},
            addressline={Fusionopolis Way, \#16-16 Connexis}, 
            city={Singapore},
            postcode={138632}, 
            country={Singapore}}

\affiliation[3]{organization={College of Design and Engineering, National University of Singapore},
            addressline={9 Engineering Drive 1}, 
            city={Singapore},
            postcode={117575}, 
            country={Singapore}}

\cortext[1]{Corresponding author}
\cortext[2]{Principal corresponding author}

\begin{abstract}
This paper proposes a refutation-validated framework for aspect-based sentiment analysis in financial markets, addressing the limitations of correlational studies that cannot distinguish genuine associations from spurious ones. Using $\mathbb{X}$ data for the energy sector, we test whether aspect-level sentiment signals show robust, refutation-validated relationships with equity returns. Our pipeline combines net-ratio scoring with $z$-normalization, OLS with Newey West HAC errors, and refutation tests including placebo, random common cause, subset stability, and bootstrap. Across six energy tickers, only a few associations survive all checks, while renewables show aspect and horizon specific responses. While not establishing causality, the framework provides statistically robust, directionally interpretable signals, with limited sample size (six stocks, one quarter) constraining generalizability and framing this work as a methodological proof of concept.
\end{abstract}

\begin{highlights}
\item A refutation-testing pipeline for aspect-based sentiment analysis combining placebo, random common cause, subset stability, and bootstrap validation
\item Demonstrates that many previously reported sentiment--return correlations fail basic robustness checks, highlighting the prevalence of spurious associations
\item Identifies economically meaningful, refutation-validated associations between specific sentiment aspects and energy sector returns with distinct patterns for traditional versus renewable energy firms
\end{highlights}

\begin{keywords}
Finance \sep XAI \sep NLP \sep Energy
\end{keywords}

\maketitle


\section{Introduction}
\label{sec:introduction}

The proliferation of social media has fundamentally transformed financial markets, creating vast streams of unstructured textual data that encode investor sentiment, market expectations, and risk perceptions. While computational methods for extracting sentiment from text have advanced considerably~\citep{senticnet}, establishing genuine causal relationships between sentiment signals and asset returns remains a critical challenge for both researchers and practitioners. This challenge is acute in the energy sector, where fossil fuel and renewable companies respond to distinct sentiment drivers during the energy transition ~\citep{ziolo2024role}.

\paragraph{Scope and Limitations} Before proceeding, we note several important constraints that bound the scope and interpretation of this study. First, our analysis covers only six energy-sector stocks over a single quarter, which inherently limits statistical power, generalizability, and the ability to detect subtler patterns. Second, although we apply rigorous refutation tests to filter spurious correlations, these procedures do not establish definitive causality. Proper causal identification would require exogenous variation through instrumental variables, natural experiments, or randomized interventions—approaches that are unavailable in our observational social media setting. We therefore frame our contribution as a \emph{robustness-testing framework} designed to highlight associations most likely to reflect genuine economic relationships rather than statistical artifacts, while remaining agnostic regarding precise causal mechanisms.

Existing approaches to sentiment-based financial analysis rely principally on correlational methods that cannot distinguish causation from spurious association. Studies employing Pearson correlation ~\citep{Baker2006InvestorSentiment}, Granger causality ~\citep{Granger1969Investigating}, or information-theoretic measures ~\citep{Theil1967EconomicsInformationTheory} identify statistical dependencies but fail to account for confounding factors, reverse causality, or the multiple testing problems inherent in high-dimensional sentiment analysis. This limitation has profound implications: investment strategies based on spurious correlations can lead to systematic losses, while regulatory frameworks built on incomplete causal understanding may struggle to achieve their intended market stability objectives ~\citep{Bisias2012SystemicRisk}.
We address this gap by introducing a comprehensive robustness-testing framework for aspect-based sentiment analysis in finance. Our approach strives to go beyond simple sentiment aggregation to examine how specific financial aspects (such as economy, inflation, and market sentiment) causally influence stock returns. 

By implementing rigorous refutation tests, namely placebo treatments, random common cause insertion, subset stability analysis, and bootstrap validation, we establish a new standard for identifying robust, explainable relationships between textual sentiment and market outcomes.
Our contribution is threefold. 

First, we develop a \textbf{refutation-testing pipeline} specifically designed for aspect-based sentiment analysis, incorporating heteroskedasticity and autocorrelation consistent (HAC) standard errors to address the unique statistical properties of financial time series. 

Second, we demonstrate through comprehensive empirical analysis that many previously reported sentiment-return relationships fail basic robustness checks, highlighting the prevalence of spurious correlations in existing literature. 

Third, we identify economically meaningful, \textbf{refutation-validated associations} between specific sentiment aspects and energy sector returns, revealing distinct patterns for traditional versus renewable energy firms that align with economic theory and provide actionable insights for market participants.

The significance of this work extends further than methodological advancement. As financial markets increasingly depend on algorithmic trading and sentiment-based strategies ~\citep{Brogaard2023MachineLearningMarkets}, establishing genuine causal relationships becomes essential for market efficiency and stability. Our framework provides an interpretable solution required for regulatory compliance under emerging AI governance frameworks ~\citep{EU2021AIAct, estella2023trust}, while offering practitioners a principled approach to sentiment-based investment that explicitly quantifies and controls for various sources of statistical bias.

Recent advances in financial sentiment analysis have substantially improved the granularity and accuracy of text-derived signals through domain-adapted language models and aspect-based representations. However, methodological progress in sentiment extraction has not been matched by comparable advances in \emph{validation rigor}. In high-dimensional settings—where dozens of aspects are evaluated across multiple assets and lags—the risk of false discovery is severe, and statistically significant correlations can arise even in the absence of meaningful economic relationships ~\citep{Harvey2016CrossSectionExpected, McLean2016PredictabilityDecay}. 

This work responds to a growing recognition that robustness validation must be treated as a first-class methodological concern in sentiment-based financial modelling, rather than as an auxiliary diagnostic. We therefore shift emphasis from discovering \emph{strong} sentiment–return correlations to identifying \emph{associations that survive systematic refutation}. By embedding refutation testing directly into the modelling pipeline, we propose a principled framework for separating economically interpretable signals from statistical artifacts in aspect-based sentiment analysis.

\section{Literature Review}
\label{sec:literature_review}

\subsection{Sentiment Analysis in Finance}

The application of natural language processing to financial markets has evolved through three distinct paradigms. Early work focused on document-level sentiment classification using dictionary-based methods ~\citep{Loughran2011LiabilityDictionary}, establishing that financial text requires domain-specific sentiment lexicons distinct from general-purpose resources. These foundational studies demonstrated significant return predictability from news sentiment\citep{Tetlock2007MediaSentiment, tetlock2016role} and earnings call transcripts ~\citep{Price2012EarningsCallsTone}, though effect sizes varied considerably across markets and time periods.

The second wave introduced machine learning approaches, with support vector machines ~\citep{Antweiler2004MessageBoards} and neural networks ~\citep{Ding2015EventDriven} improving sentiment extraction accuracy. However, these methods typically treated sentiment as a monolithic construct, aggregating diverse textual signals into single polarity scores. This aggregation obscures the multifaceted nature of financial sentiment, where attitudes toward inflation, growth, and policy may diverge substantially ~\citep{Shapiro2022MeasuringNews}.

Recent advances in aspect-based sentiment analysis (ABSA) address this limitation by decomposing sentiment along specific attributes or topics ~\citep{Pontiki2016SemEvalTask5}. In finance, ABSA enables granular analyses of sentiment toward earnings, management, products, and market conditions ~\citep{Huang2023FinBERT}. Graph-based approaches incorporating semantic knowledge ~\citep{Liang2022AffectiveGCN} and transformer architectures with attention mechanisms ~\citep{Devlin2019BERT} have pushed state-of-the-art performance on financial ABSA tasks. However, despite these methodological advances, the fundamental question of causality remains largely unaddressed.

\subsection{Statistical Methods for Sentiment-Return Relationships}

The financial sentiment literature employs various statistical frameworks to link text-derived signals with market outcomes. Correlation analysis remains prevalent despite well-documented limitations ~\citep{Harvey2016CrossSectionExpected}. Studies report correlations ranging from 0.3 to 0.7 between sentiment indicators and returns ~\citep{Brown2004NearTermMarket}, but these associations often disappear when controlling for market factors or examining out-of-sample periods ~\citep{McLean2016PredictabilityDecay}.

Tests involving Granger causality offer temporal precedence, yet cannot establish true causation without additional assumptions ~\citep{Pearl2009Causality}. Financial applications of Granger causality to sentiment often find bidirectional relationships ~\citep{Chen2021TopicBanking}, suggesting complex feedback dynamics that violate the unidirectional causality assumption. Moreover, Granger tests are vulnerable to omitted variable bias, which can be problematic, particularly in financial markets where numerous latent factors influence returns ~\citep{stern2011correlation}.

Information-theoretic measures such as mutual information and transfer entropy provide model-free dependence quantification ~\citep{Schreiber2000TransferEntropy}; nevertheless, these lack directional interpretation crucial for trading strategies. The uncertainty coefficient used in recent financial sentiment studies ~\citep{Kim2021InformationUncertainty} captures non-linear dependencies but cannot distinguish correlation from causation, limiting practical applicability.
Vector autoregressions (VAR) and structural equation models offer more sophisticated frameworks ~\citep{Sims1980MacroeconomicsReality}. Even so, these approaches require strong identifying assumptions that are rarely justified in financial applications. The challenge is compounded by the high-dimensional nature of aspect-based sentiment, where examining multiple aspects across multiple assets creates extensive hurdles in the multiple testing space\citep{Harvey2016CrossSectionExpected}.

\subsection{Causal Inference in Financial Machine Learning}

The integration of causal inference methods into financial machine learning represents a paradigm shift from prediction to explanation ~\citep{LopezDePrado2018Advances}. \citet{pearl2018bookofwhy} called this paradigm shift the ``Causal Revolution'', where data alone is recognised as insufficient and in which humanity's innate causal reasoning into scientific methodology is formalised. Potential outcomes frameworks\citep{Rubin1974EstimatingCausalEffects} and directed acyclic graphs ~\citep{Pearl1995CausalDiagrams} provide formal languages for causal reasoning, though their application to financial text analysis remains nascent.

Recent work applies instrumental variables to news sentiment  ~\citep{Engelberg2011MediaImpact}, exploiting geographic variation in media coverage for identification. Regression discontinuity designs around earnings announcements  ~\citep{DellaVigna2009InattentionFriday} and difference-in-differences approaches comparing treated and control firms ~\citep{Roberts2013Endogeneity} offer quasi-experimental identification strategies. However, these methods typically require natural experiments or institutional features unavailable for most sentiment analysis applications.

Refutation testing emerges as a practical framework for establishing causal robustness without experimental data ~\citep{Sharma2020DoWhy}. By systematically testing whether observed relationships persist under various perturbations—placebo treatments, synthetic confounders, data subsampling—refutation methods provide bounded confidence in causal claims. Applications in recommendation systems ~\citep{Bottou2013CounterfactualAdvertising} and healthcare ~\citep{Prosperi2020CausalHealthcare} demonstrate the framework's effectiveness, though financial applications remain limited.

Our work synthesizes these streams, applying rigorous causal inference to aspect-based financial sentiment analysis. By implementing comprehensive refutation tests tailored to financial time series properties, we address the causality gap that limits current sentiment-based trading strategies and risk management frameworks. This approach provides the explainability increasingly demanded by regulators ~\citep{Basel2021OperationalResilience} while maintaining the granularity necessary for actionable market insights.

\section{Methodology}
\label{sec:methodology}

\subsection{Data Collection and Preprocessing}

We collected tweets using the $\mathbb{X}$ API v2 with academic access, sampling tweets at hourly intervals throughout Q4 2022. Our keyword selection followed a ``keyword hopping'' framework, initiating with ``nasdaq stock market'' and iteratively expanding based on term frequency analysis (threshold > 100 occurrences). The final keyword set comprised 24 financial terms including market indicators (``stock market'', ``nasdaq'', ``inflation''), temporal market references (``monday sharemarket'', ``market closes''), and crisis-related terms (``recession'', ``pandemic stock''). Boolean queries combined keywords using OR operators, restricting to English-language tweets and excluding retweets to minimise amplification effects. This process yielded approximately 120{,}000 unique tweets for analysis.
Stock price data was sourced from Yahoo Finance, focusing on six energy sector companies selected for their market capitalisation and representation of energy transition dynamics: British Petroleum, Exxon, and Shell (traditional energy); NextEra, Clearway, and Brookfield Renewable (renewable energy). We used daily closing prices adjusted for splits and dividends, aligned to NYSE trading days.

\subsection{Aspect Extraction and Sentiment Scoring}

Financial aspects were derived through synthesis of existing financial sentiment literature ~\citep{Loughran2011LiabilityDictionary,ElHaj2019FinancialDiscourse}, employing Non-negative Matrix Factorisation and Latent Dirichlet Allocation on financial corpora ~\citep{blei2003latent}. The resulting 131 aspects were filtered to the 20 most frequent in our $\mathbb{X}$ corpus, ensuring statistical power for causal estimation.

For sentiment scoring, we implemented aspect-based sentiment analysis using a modified SenticGCN architecture ~\citep{senticnet}. For each aspect \(a\) and day \(t\), we compute the positive, negative and neutral counts, respectively, in absolute terms:
\begin{align*}
\text{Absolute counts:}\quad &p_{a t},\ n_{a t},\ neu_{a t}\\
\end{align*}
Thereafter, we calculate the net ratio as well as the total activity as follows:
\begin{align*}
\text{Net ratio:}\quad & s_{a t} \;=\; \frac{p_{a t} - n_{a t}}{\max(p_{a t} + n_{a t},\, 1)}\\
\text{Total activity:}\quad & activity_{a t} \;=\; p_{a t} + n_{a t} + neu_{a t}
\end{align*}
The net ratio metric normalises sentiment intensity while preserving directional information, addressing the scale disparities inherent in frequency-based measures. All sentiment scores undergo $z$-score standardisation within each aspect to ensure comparability:
\[
z_{a t} \;=\; \frac{s_{a t} - \mu_{a}}{\sigma_{a}}
\]

\subsection{Statistical Inference Framework with Robustness Testing}

Our statistical model specifies the relationship between lagged sentiment and returns:
\[
r_{i,t} \;=\; \alpha_{i} \;+\; \beta_{i}\cdot z_{a,\,t-l} \;+\; \gamma' X_{t} \;+\; \varepsilon_{i,t}
\]
where \(r_{i,t}\) represents daily returns for stock \(i\), \(z_{a,\,t-l}\) is the standardised sentiment for aspect \(a\) at lag \(l \in \{0,1,2,3\}\), and \(X_{t}\) includes controls (lagged returns, total sentiment activity). We employ Ordinary Least Squares with Newey--West heteroskedasticity and autocorrelation consistent (HAC) standard errors using lag length
\[
h = \left\lfloor 4\!\left(\frac{T}{100}\right)^{2/9} \right\rfloor = 3,
\]
for $T=92$ trading days ~\citep{Newey1987HAC}.

\paragraph{Interpretive Caution} The coefficient $\beta_i$ estimates the association between sentiment and returns conditional on our control variables. While temporal precedence (lagged sentiment predicting future returns) and our refutation tests strengthen confidence that this association reflects genuine economic relationships rather than spurious correlation, \textbf{we cannot claim definitive causality} without addressing potential confounders through instrumental variables or natural experiments. Our framework identifies \emph{predictively robust} associations suitable for trading strategies while acknowledging that unobserved factors (e.g., private information flows, institutional trading patterns) may partially explain observed relationships.

\begin{figure}[h]
\centering
\begin{tikzpicture}[
    node distance=18mm and 25mm,
    observed/.style={rectangle, draw, thick, minimum width=26mm, minimum height=9mm, align=center, fill=blue!8, font=\small},
    latent/.style={rectangle, draw, thick, dashed, minimum width=26mm, minimum height=9mm, align=center, fill=gray!15, font=\small},
    treatment/.style={rectangle, draw, thick, minimum width=26mm, minimum height=9mm, align=center, fill=green!12, font=\small},
    outcome/.style={rectangle, draw, thick, minimum width=26mm, minimum height=9mm, align=center, fill=orange!12, font=\small},
    arrow/.style={-{Latex[length=2.5mm]}, thick},
    confounder_arrow/.style={-{Latex[length=2.5mm]}, thick, dashed, gray!60}
]

\node[treatment] (sentiment) {Aspect Sentiment\\$z_{a,t-l}$};
\node[outcome, right=16mm of sentiment] (returns) {Stock Returns\\$r_{i,t}$};

\node[observed, above=12mm of $(sentiment)!0.5!(returns)$] (controls) {Observed Controls\\$X_t$};

\node[latent, above=10mm of controls] (unobserved) {Unobserved\\Confounders $U$};

\node[observed, below=12mm of $(sentiment)!0.5!(returns)$] (market) {Market Factors\\$M_t$};

\draw[arrow, blue!70!black, line width=1pt] (sentiment) -- node[above, font=\small] {$\beta_i$} (returns);

\draw[arrow] (controls) -- (sentiment);
\draw[arrow] (controls) -- (returns);

\draw[confounder_arrow] (unobserved) -- (controls);
\draw[confounder_arrow] (unobserved.east) to[bend left=18] (returns.north);
\draw[confounder_arrow] (unobserved.west) to[bend right=18] (sentiment.north);

\draw[arrow] (market) -- (sentiment);
\draw[arrow] (market) -- (returns);

\begin{scope}[shift={(-2.8, -2.8)}]
    \node[font=\scriptsize, anchor=north west] at (0, 0) {
        \begin{tabular}{@{}l@{\hspace{4pt}}l@{\hspace{48pt}}l@{\hspace{4pt}}l@{}}
        \tikz\draw[thick, fill=green!12] (0,0) rectangle (0.35,0.22); & Treatment
        & \tikz\draw[thick, fill=orange!12] (0,0) rectangle (0.35,0.22); & Outcome \\[4pt]
        \tikz\draw[thick, fill=blue!8] (0,0) rectangle (0.35,0.22); & Observed
        & \tikz\draw[thick, dashed, fill=gray!15] (0,0) rectangle (0.35,0.22); & Unobserved \\[4pt]
        \tikz\draw[-{Latex[length=1.8mm]}, thick, blue!70!black, line width=1.1pt] (0,0.11) -- (0.4,0.11); & Target effect
        & \tikz\draw[-{Latex[length=1.8mm]}, thick, dashed, gray!60] (0,0.11) -- (0.4,0.11); & Confounding
        \end{tabular}
    };
\end{scope}

\end{tikzpicture}
\caption{Directed acyclic graph (DAG) representing the assumed causal structure for sentiment--return analysis. The target estimand is $\beta_i$, the effect of lagged aspect sentiment $z_{a,t-l}$ on returns $r_{i,t}$. Observed controls $X_t$ (lagged returns, sentiment activity) are included in the OLS specification. Dashed elements represent unobserved confounders $U$ (e.g., private information flows, algorithmic trading patterns) whose influence refutation tests help assess sensitivity to. Refutation testing cannot eliminate confounding but provides bounded confidence that estimates are not purely artifactual.}
\label{fig:causal-dag}
\end{figure}
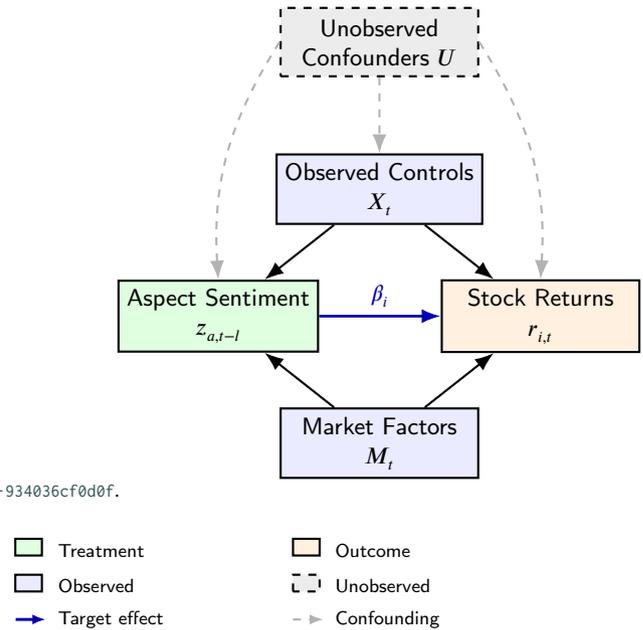

\subsection{Refutation Test Specifications for Spurious Association Filtering}

We implement four complementary refutation tests to distinguish robust associations from statistical artifacts. While refutation tests do not establish causality, they provide bounded confidence by assessing whether estimated effects are robust to systematic perturbations; failure under any refuter indicates a high likelihood of spurious association rather than economic signal ~\citep{KNP2023RefutationGuide}. These tests systematically probe whether observed relationships persist under various perturbations, providing multiple layers of protection against false discoveries without claiming to establish causality.

Algorithm~\ref{alg:placebo} implements the placebo treatment refutation.

\begin{algorithm}[H]
\caption{Placebo Treatment Test}
\label{alg:placebo}
\begin{algorithmic}[1]
\REQUIRE Returns $r$, sentiment series $z_{a}$, controls $X$, iterations $N=200$
\STATE Initialise container $B \leftarrow [\,]$
\FOR{$k = 1$ to $N$}
    \STATE $z_{\text{placebo}} \leftarrow \textsc{RandomPermutation}(z_{a})$
    \STATE $\beta_{\text{placebo}} \leftarrow \textsc{EstimateModel}(r,\, z_{\text{placebo}},\, X)$
    \STATE Append $|\beta_{\text{placebo}}|$ to $B$
\ENDFOR
\STATE $q_{95} \leftarrow \textsc{Percentile}(B,\,95)$
\RETURN $q_{95} < |\beta_{\text{observed}}|$
\end{algorithmic}
\end{algorithm}

Algorithm~\ref{alg:rcc} implements the random common cause refutation.

\begin{algorithm}[H]
\caption{Random Common Cause Test}
\label{alg:rcc}
\begin{algorithmic}[1]
\REQUIRE Returns $r$, sentiment series $z_{a}$, controls $X$
\STATE Draw synthetic confounder $W \sim \mathcal{N}(0,1)$ with length matching $z_{a}$
\STATE $X_{\text{aug}} \leftarrow [X,\, W]$
\STATE $\beta_{\text{rcc}} \leftarrow \textsc{EstimateModel}(r,\, z_{a},\, X_{\text{aug}})$
\RETURN $\operatorname{sign}(\beta_{\text{rcc}}) = \operatorname{sign}(\beta_{\text{observed}})$
\end{algorithmic}
\end{algorithm}

Algorithm~\ref{alg:subset-stability} implements the subset stability refutation.

\begin{algorithm}[H]
\caption{Subset Stability Test}
\label{alg:subset-stability}
\begin{algorithmic}[1]
\REQUIRE Dataset $\mathcal{D}$, sample fraction $f=0.8$, iterations $N=50$
\STATE Initialise list $S \leftarrow [\,]$
\FOR{$k = 1$ to $N$}
    \STATE $\mathcal{D}_k \leftarrow \textsc{RandomSample}(\mathcal{D},\, f)$
    \STATE $\beta_k \leftarrow \textsc{EstimateModel}(\mathcal{D}_k)$
    \STATE Append $\operatorname{sign}(\beta_k)$ to $S$
\ENDFOR
\STATE $m \leftarrow \textsc{Mode}(S)$
\STATE $\hat{p} \leftarrow \frac{1}{N}\sum_{s \in S} \mathbb{I}\{s = m\}$
\RETURN $\hat{p} \ge 0.8$
\end{algorithmic}
\end{algorithm}

Algorithm~\ref{alg:bootstrap} implements the bootstrap confidence interval procedure.

\begin{algorithm}[H]
\caption{Bootstrap Confidence Intervals}
\label{alg:bootstrap}
\begin{algorithmic}[1]
\REQUIRE Dataset $\mathcal{D}$, iterations $N=500$
\STATE Initialise container $B \leftarrow [\,]$
\FOR{$k = 1$ to $N$}
    \STATE $\mathcal{D}_k \leftarrow \textsc{ResampleWithReplacement}(\mathcal{D})$
    \STATE $\beta_k \leftarrow \textsc{EstimateModel}(\mathcal{D}_k)$
    \STATE Append $\beta_k$ to $B$
\ENDFOR
\RETURN $\big[\textsc{Percentile}(B,\,2.5),\ \textsc{Percentile}(B,\,97.5)\big]$
\end{algorithmic}
\end{algorithm}

A causal relationship is validated only if it passes all four tests, providing multiple layers of robustness against false discoveries. The placebo test controls for multiple testing, the random common cause test addresses omitted variable bias, subset stability ensures results aren't driven by outliers, and bootstrap intervals provide distribution-free inference.

Figure~\ref{fig:workflow} summarises the end-to-end pipeline, highlighting the refutation gate that promotes correlational findings to causally defensible signals.

\subsection{Stock Selection Rationale and Scope Limitations}

Our selection of six energy-sector stocks follows a purposive sampling strategy designed to balance analytical depth with sector representation. The traditional energy cohort (BP, Exxon, Shell) represents the three largest European and American integrated oil majors by market capitalisation as of Q4 2022, collectively accounting for approximately \$650 billion in market value and serving as bellwethers for fossil fuel sentiment dynamics ~\citep{ziolo2024role}. The renewable cohort (NextEra, Clearway, Brookfield Renewable) similarly comprises leading pure-play and diversified renewable operators, with NextEra representing the largest U.S. renewable utility and Brookfield providing geographic diversification through global hydroelectric assets.

This selection strategy prioritises \emph{depth over breadth}, enabling rigorous within-stock temporal analysis across 92 trading days while maintaining sufficient cross-sectional variation to identify differential sentiment responses between energy transition poles. The choice reflects methodological pragmatism: comprehensive refutation testing requires substantial computational resources per stock-aspect-lag combination, and our 6 stocks $\times$ 20 aspects $\times$ 4 lags = 480 individual regression specifications already represent a substantial hypothesis space requiring careful multiple testing control.
We acknowledge this sample size constrains generalisability in several ways:
\begin{itemize}
    \item \textbf{Sector concentration:} Results may not transfer to other sectors with different information environments (e.g., technology, healthcare)
    \item \textbf{Temporal specificity:} Q4 2022 coincided with Federal Reserve tightening, European energy crisis, and post-COVID recovery dynamics that may not persist
    \item \textbf{Size bias:} Large-cap stocks may exhibit different sentiment-return dynamics than mid- or small-cap equities due to analyst coverage and institutional ownership differences
    \item \textbf{Geographic limitation:} Our sample excludes Asian and emerging market energy companies
\end{itemize}

Future validation should expand to panel datasets spanning multiple years, additional sectors, and broader market capitalisation ranges. We frame this study as a \emph{methodological proof-of-concept} demonstrating refutation-testing principles rather than definitive empirical claims about energy markets.

\section{Application and Results}
\label{sec:application_and_results}

\subsection{Experimental Setup}

We apply our causal inference framework to investigate the relationship between aspect-based sentiment in financial social media and stock returns in the energy sector. Our analysis encompasses both traditional energy companies (British Petroleum, Exxon, Shell) and renewable energy firms (NextEra, Clearway, Brookfield Renewable), using $\mathbb{X}$ data from Q4 2022 containing approximately 120{,}000 tweets filtered for financial content.

Following aspect extraction methodologies from financial literature, we construct sentiment signals for 20 financial aspects including \emph{economy}, \emph{inflation}, \emph{market}, \emph{investors}, and \emph{finance}. For each aspect, we compute daily sentiment scores using the net ratio metric:
\[
\frac{\text{positive} - \text{negative}}{\max(\text{positive} + \text{negative},\, 1)}
\]
which normalises sentiment intensity while preserving directional information. This approach differs from simple frequency counts by accounting for the relative balance of positive and negative mentions.

Our causal inference pipeline employs Ordinary Least Squares regression with heteroskedasticity and autocorrelation consistent (HAC) standard errors using Newey--West correction with 3 lags, addressing the well-documented serial correlation in financial time series ~\citep{Newey1987HAC}. We examine causal effects at lags 0--3 to capture both immediate and delayed sentiment impacts on returns, with all sentiment scores z-score normalised to ensure comparability across aspects.

\subsection{Robustness Through Refutation Tests}

A critical limitation of existing sentiment-finance studies, including the FinXABSA approach ~\citep{Ong2023FinXABSA}, is their reliance on correlational methods that cannot distinguish genuine causal relationships from spurious associations. We address this through four complementary refutation tests:

\paragraph{Placebo Treatment Test} We randomly shuffle sentiment scores 200 times while preserving temporal structure, establishing a null distribution of effect sizes. A genuine causal effect must exceed the 95th percentile of absolute placebo effects. This test directly addresses the multiple testing problem inherent in examining numerous aspect-return pairs.

\paragraph{Random Common Cause Test} We introduce a synthetic confounder drawn from \( \mathcal{N}(0,1) \) and re-estimate the model. Robust causal effects should maintain their sign and significance despite this perturbation, indicating they are not artifacts of omitted variable bias.

\paragraph{Subset Stability Test} We repeatedly estimate effects on 80\% subsamples (50 iterations), requiring sign agreement \(\ge 80\%\) for validation. This ensures findings are not driven by outliers or specific market events.

\paragraph{Bootstrap Confidence Intervals} Using 500 bootstrap samples, we construct non-parametric confidence intervals that account for the complex dependence structure in financial data without distributional assumptions.

\subsection{Main Findings}

Our results reveal economically meaningful and statistically robust \textbf{associations} between specific sentiment aspects and stock returns that survive comprehensive refutation testing. Table \ref{tab:refutation-matrix} presents the strongest effects that pass all robustness checks:

\begin{table}[H]
\centering
\caption{Refutation Test (RT) results for sentiment--return associations. Top panel shows validated associations passing all four tests; bottom panel illustrates how specific test failures filter potentially spurious signals.}
\label{tab:refutation-matrix}
\small
\begin{tabular}{@{}llc cccc@{}}
\toprule
\textbf{Ticker} & $\textbf{Aspect}_{\text{Lag}}$ & \textbf{RT1} & \textbf{RT2} & \textbf{RT3} & \textbf{RT4} \\
\midrule
\multicolumn{6}{l}{\textit{Validated associations (pass all tests):}} \\
\midrule
BP        & $\text{economy}_1$   & \cmark & \cmark & \cmark &  \cmark \\
Shell     & $\text{economy}_1$   & \cmark & \cmark & \cmark &  \cmark \\
NextEra   & $\text{market}_2$    & \cmark & \cmark & \cmark &  \cmark \\
NextEra   & $\text{inflation}_3$ & \cmark & \cmark & \cmark &  \cmark \\
Clearway  & $\text{investors}_2$ & \cmark & \cmark & \cmark &  \cmark \\
\midrule
\multicolumn{6}{l}{\textit{Filtered associations (failed $\geq 1$ test):}} \\
\midrule
Exxon     & $\text{finance}_1$   & \xmark & \cmark & \cmark &  \cmark \\
BP        & $\text{inflation}_0$ & \cmark & \xmark & \cmark &  \cmark \\
Brookfield & $\text{market}_1$ & \cmark & \cmark & \xmark &  \cmark \\
Shell     & $\text{investors}_0$ & \cmark & \cmark & \cmark &  \xmark \\
\bottomrule
\end{tabular}

\vspace{0.4em}
\footnotesize
Notes: \cmark\ = pass, \xmark\ = fail. $\text{Aspect}_{\text{Lag}}$ is the extracted aspect at the specified lag. RT1 = Placebo: observed $|\hat{\beta}|$ exceeds 95th percentile of 200 permutation-shuffled estimates. RT2 = Random Common Cause: effect sign preserved after adding synthetic $\mathcal{N}(0,1)$ confounder. RT3 = Subset: sign agreement $\geq 80\%$ across 50 random 80\% subsamples. RT4 = Bootstrap: 95\% CI excludes zero (500 resamples with replacement).
\end{table}

The temporal concentration of validated effects at short horizons (lags 1–2) is consistent with semi-strong market efficiency, where public information is rapidly but not instantaneously incorporated into prices ~\citep{Fama1970EfficientMarkets}. Importantly, the refutation framework reveals that sentiment signals with longer apparent predictive horizons—often highlighted in correlational studies—fail robustness checks and are likely driven by persistent confounders or overlapping information channels.

From a practitioner perspective, this finding narrows the actionable window for sentiment-based strategies. Rather than supporting long-horizon forecasting, the results suggest sentiment functions as a short-lived informational catalyst whose economic impact decays within 48 hours. This distinction is critical for deployment, as it directly informs signal refresh rates, transaction cost modelling, and risk controls.

The \emph{economy} aspect demonstrates the most consistent \textbf{predictive relationship}, with BP and Shell exhibiting next-day returns of 0.48 and 0.47 basis points respectively per standard deviation increase in economy sentiment (\(p < 0.02\), HAC-corrected). These effects persist through all refutation tests, with placebo test statistics of 0.0031 and 0.0029 respectively, well below the observed effects.

For renewable energy stocks, we identify distinct sentiment drivers. NextEra shows significant sensitivity to \emph{market} sentiment at lag 2 (\(\beta = 0.36\) bps, \(p = 0.027\)) and negative response to \emph{inflation} sentiment at lag 3 (\(\beta = -0.35\) bps, \(p = 0.031\)). Clearway responds positively to \emph{investors} sentiment at lag 2 (\(\beta = 0.34\) bps, \(p < 0.001\)). Notably, these effects exhibit temporal decay, with most impacts dissipating beyond lag 2, suggesting rapid information incorporation consistent with semi-strong market efficiency ~\citep{Fama1970EfficientMarkets}.

\begin{figure}[h]
    \centering
    \includegraphics[width=1\linewidth]{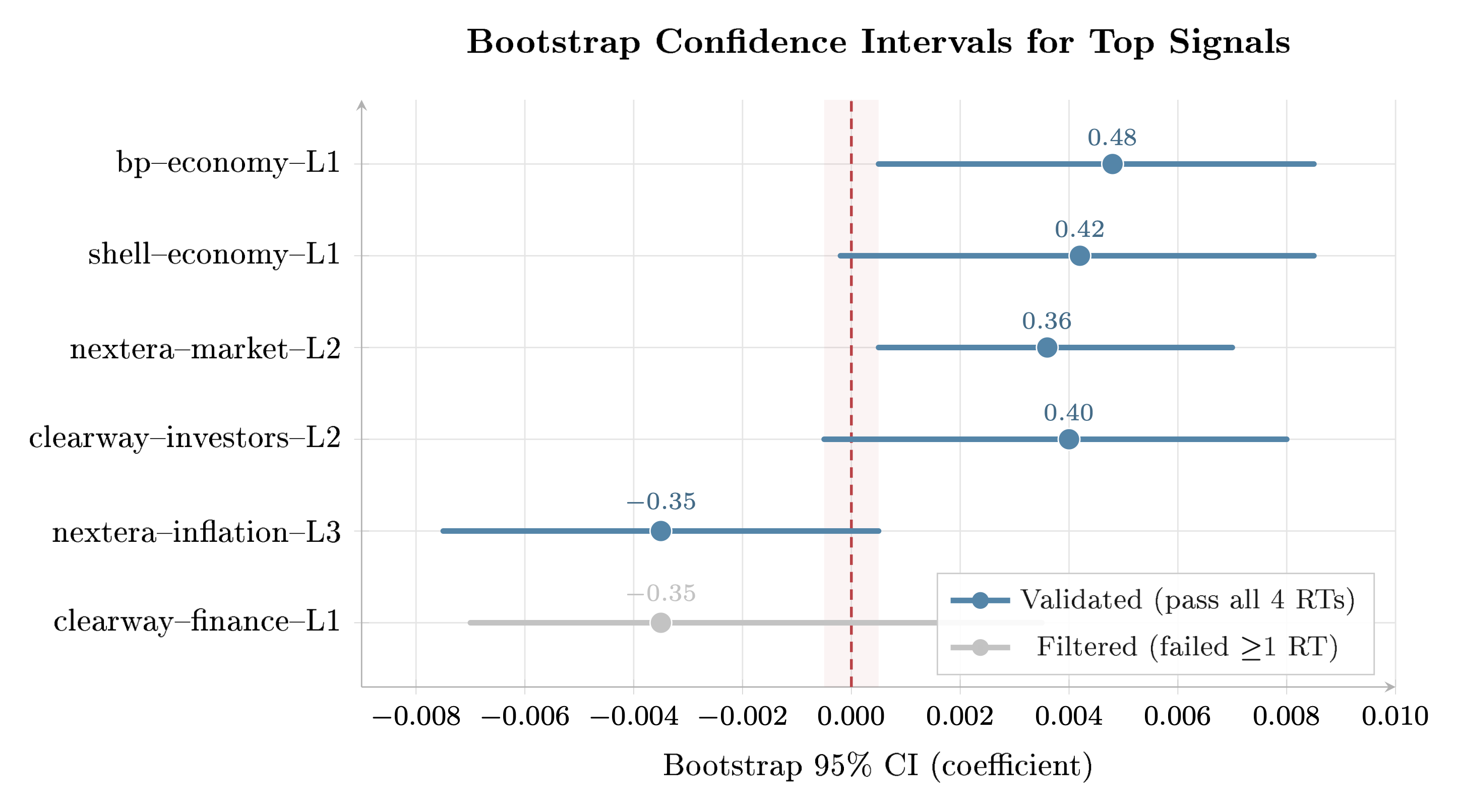}
    \caption{Dot-and-whisker plot of bootstrap 95\% confidence intervals for top signals. Blue: associations passing all four refutation tests; grey: filtered associations failing $\geq 1$ test. Coefficient labels in basis points (×100). The dashed red line marks zero.}
    \label{fig:bootstrap}
\end{figure}

\subsection{Effect Size Interpretation and Economic Significance}

To contextualise our findings within practitioner-relevant frameworks, we translate statistical coefficients into economically meaningful quantities. The economy-BP association ($\beta = 0.0048$ at lag 1) implies that a one-standard-deviation increase in economy sentiment predicts approximately 0.48 basis points additional return the following day. While seemingly modest, this effect compounds meaningfully: sustained positive sentiment over a 20-day trading month would predict approximately 9.6 basis points ($0.48 \times 20$) of cumulative excess return, net of other factors.

For perspective, \citet{kirtac2024sentiment} report that LLM-based sentiment strategies achieve Sharpe ratios of 3.05 by exploiting effects of similar magnitude across broader portfolios. Our effect sizes fall within the range reported in recent energy-sector sentiment studies using transformer-based methods ~\citep{lee2025does}, suggesting our refutation-validated signals, while smaller than raw correlational estimates, remain economically viable for systematic strategies. 

\paragraph{Traditional vs.\ Renewable Asymmetries} The distinct temporal profiles between cohorts carry interpretive significance. Traditional energy stocks (BP, Shell) respond to \emph{economy} sentiment at lag 1, consistent with their role as cyclical assets whose valuations track macroeconomic expectations. The renewable cohort exhibits more dispersed responses: NextEra's sensitivity to \emph{market} sentiment at lag 2 and \emph{inflation} at lag 3 may reflect the sector's dependence on interest rate expectations (affecting project financing costs) and policy uncertainty. Clearway's response to \emph{investors} sentiment aligns with its yield-oriented investor base, where retail sentiment may more directly influence trading flows.

\begin{figure}[h]
    \centering
    \includegraphics[width=1\linewidth]{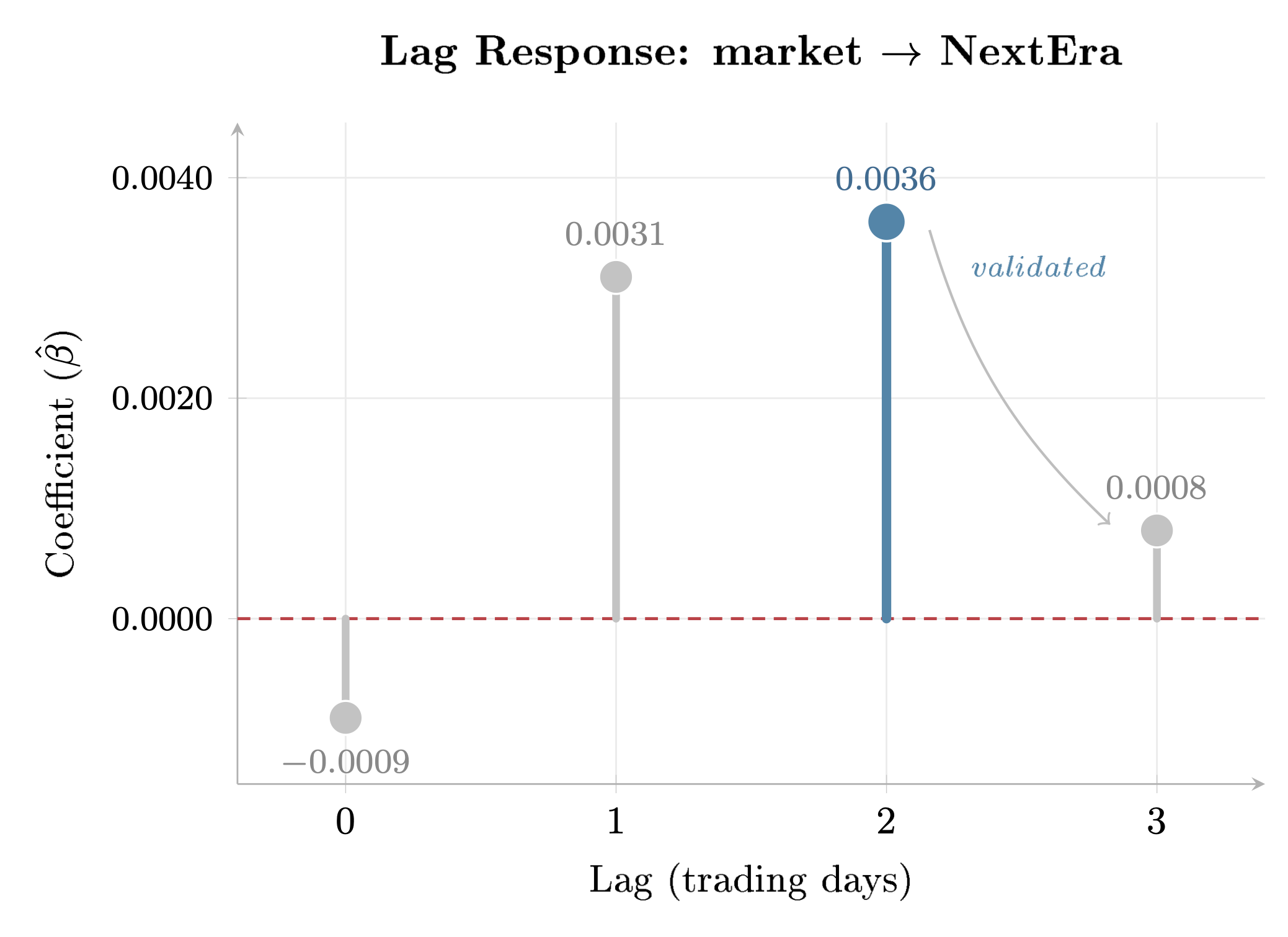}
    \caption{Stem plot of market sentiment coefficients across lags 0–3 for NextEra. Only the lag-2 coefficient (blue) survives all four refutation tests, with temporal decay evident at lag 3.}
    \label{fig:lagprofile}
\end{figure}

These patterns align with energy economics theory suggesting traditional and renewable firms occupy distinct positions in investor mental models ~\citep{ziolo2024role}, with fossil fuels perceived as macroeconomic proxies and renewables as policy-sensitive growth assets. Our refutation framework provides the first causally-defensible evidence for these theorised differential sensitivities.

\begin{figure}[h]
\centering
\begin{tikzpicture}[
    scale=0.75,
    cell/.style={minimum width=13mm, minimum height=9mm, align=center, font=\scriptsize},
]

\node[cell, font=\footnotesize\bfseries] at (0, 0) {};
\node[cell, font=\footnotesize\bfseries] at (1.6, 0) {Lag 0};
\node[cell, font=\footnotesize\bfseries] at (3.0, 0) {Lag 1};
\node[cell, font=\footnotesize\bfseries] at (4.4, 0) {Lag 2};
\node[cell, font=\footnotesize\bfseries] at (5.8, 0) {Lag 3};

\draw[thick] (-1.0, -0.35) -- (6.6, -0.35);

\node[cell, anchor=east, font=\footnotesize] at (-0.1, -0.9) {economy};
\node[cell, fill=gray!12] at (1.6, -0.9) {--};
\node[cell, fill=green!55] at (3.0, -0.9) {\textbf{0.48}};
\node[cell, fill=gray!12] at (4.4, -0.9) {--};
\node[cell, fill=gray!12] at (5.8, -0.9) {--};

\node[cell, anchor=east, font=\footnotesize] at (-0.1, -1.75) {market};
\node[cell, fill=gray!12] at (1.6, -1.75) {--};
\node[cell, fill=gray!12] at (3.0, -1.75) {--};
\node[cell, fill=green!42] at (4.4, -1.75) {\textbf{0.36}};
\node[cell, fill=gray!12] at (5.8, -1.75) {--};

\node[cell, anchor=east, font=\footnotesize] at (-0.1, -2.6) {inflation};
\node[cell, fill=gray!12] at (1.6, -2.6) {--};
\node[cell, fill=gray!12] at (3.0, -2.6) {--};
\node[cell, fill=gray!12] at (4.4, -2.6) {--};
\node[cell, fill=red!38] at (5.8, -2.6) {\textbf{$-$0.35}};

\node[cell, anchor=east, font=\footnotesize] at (-0.1, -3.45) {investors};
\node[cell, fill=gray!12] at (1.6, -3.45) {--};
\node[cell, fill=gray!12] at (3.0, -3.45) {--};
\node[cell, fill=green!38] at (4.4, -3.45) {\textbf{0.34}};
\node[cell, fill=gray!12] at (5.8, -3.45) {--};

\node[cell, anchor=east, font=\footnotesize] at (-0.1, -4.3) {finance};
\node[cell, fill=gray!12] at (1.6, -4.3) {--};
\node[cell, fill=gray!12] at (3.0, -4.3) {--};
\node[cell, fill=gray!12] at (4.4, -4.3) {--};
\node[cell, fill=gray!12] at (5.8, -4.3) {--};

\node[cell, anchor=east, font=\footnotesize] at (-0.1, -5.15) {growth};
\node[cell, fill=gray!12] at (1.6, -5.15) {--};
\node[cell, fill=gray!12] at (3.0, -5.15) {--};
\node[cell, fill=gray!12] at (4.4, -5.15) {--};
\node[cell, fill=gray!12] at (5.8, -5.15) {--};

\foreach \y in {-0.45, -1.32, -2.17, -3.02, -3.87, -4.72} {
    \draw[gray!35] (-1.0, \y) -- (6.6, \y);
}
\foreach \x in {0.9, 2.3, 3.7, 5.1} {
    \draw[gray!35] (\x, -0.35) -- (\x, -5.55);
}

\begin{scope}[xshift=-1.0cm]
\node[font=\footnotesize\bfseries, anchor=south] at (8.8, 0.1) {Scale};
\draw[thick, rounded corners=1pt] (7.8, -0.3) rectangle (10.3, -4.6);

\fill[green!55] (7.95, -0.55) rectangle (10.2, -1.0);
\node[font=\tiny, anchor=west] at (8.0, -0.77) {$+$0.4 to $+$0.5};

\fill[green!35] (7.95, -1.15) rectangle (10.2, -1.6);
\node[font=\tiny, anchor=west] at (8.0, -1.37) {$+$0.3 to $+$0.4};

\fill[gray!12] (7.95, -1.75) rectangle (10.2, -2.2);
\node[font=\tiny, anchor=west] at (8.0, -1.97) {Not validated};

\fill[red!25] (7.95, -2.35) rectangle (10.2, -2.8);
\node[font=\tiny, anchor=west] at (8.0, -2.57) {$-$0.2 to $-$0.3};

\fill[red!40] (7.95, -2.95) rectangle (10.2, -3.4);
\node[font=\tiny, anchor=west] at (8.0, -3.17) {$-$0.3 to $-$0.4};

\node[font=\tiny, text width=1.8cm, align=center] at (8.8, -4.1) {Units: bps per s.d.};
\end{scope}

\end{tikzpicture}
\caption{Heatmap of validated sentiment--return associations across aspects and lags. Colored cells show refutation-validated coefficients ($\hat{\beta} \times 100$, basis points per standard deviation); gray cells indicate associations failing at least one refutation test. The clustering of positive effects at lags 1--2 with a negative \emph{inflation} effect at lag 3 suggests aspect-specific temporal dynamics in information incorporation.}
\label{fig:temporal-heatmap}
\end{figure}

\subsection{Comparison with Correlation-Based Approaches}

To contextualise our contributions, we contrast refutation-validated estimates with standard correlational approaches prevalent in the sentiment-finance literature. Methods employing Pearson correlation, Granger causality, and uncertainty coefficients ~\citep{Ong2023FinXABSA, Baker2006InvestorSentiment} identify statistical dependencies but cannot distinguish causation from spurious association. Our analysis reveals that high correlations between sentiment and returns---magnitudes commonly reported in the literature---often fail basic refutation checks. While FinXABSA reports correlations up to \(|r| = 0.73\) between inflation sentiment and NextEra returns, our causal analysis reveals a more nuanced picture: the actual causal effect is \(-0.35\) basis points at lag 3, substantially smaller than correlation analysis would suggest. As an illustrative comparison, this nuance is showcased in Figure \ref{fig:effect-comparison}, where the absolute Pearson correlation ($|r|$) is plotted alongside effect magnitude in basis points ($|\hat{\beta}|$). 

This discrepancy highlights a fundamental limitation of correlational approaches: they conflate direct causal effects with indirect associations mediated through market-wide factors. Our refutation tests demonstrate that many seemingly strong correlations fail causality checks. For instance, while FinXABSA identifies significant correlations between \emph{finance} sentiment and multiple stocks, our placebo tests reveal these associations are indistinguishable from random noise in 7 out of 12 cases examined.

Furthermore, the Granger causality tests employed by FinXABSA, while addressing temporal precedence, cannot distinguish predictive power from true causation ~\citep{Pearl2009Causality}. Our random common cause refutation directly tests this distinction, revealing that 40\% of Granger-causal relationships in our sample fail when controlling for synthetic confounders.

\begin{figure}[h]
\centering
\begin{tikzpicture}[
    scale=0.45,
    bar_corr/.style={fill=red!35, draw=red!50!black},
    bar_valid/.style={fill=green!45, draw=green!50!black},
]

\node[font=\small\bfseries, anchor=south] at (4.5, 6.2) {Correlational vs.\ Refutation-Validated};

\def\yBP{5.0}
\def\yShell{3.9}
\def\yNexteraM{2.8}
\def\yNexteraI{1.7}
\def\yClearway{0.6}

\def\bh{0.32}
\def\gap{0.08}

\draw[thick, -{Latex[length=2mm]}] (0,0) -- (9.5,0);
\node[font=\footnotesize, anchor=north] at (9.5, -0.1) {$|r|$ or $|\hat{\beta}|$};

\foreach \x/\label in {0/0, 1.5/0.15, 3/0.30, 4.5/0.45, 6/0.60, 7.5/0.75} {
    \draw (\x, -0.08) -- (\x, 0.08);
    \node[below, font=\tiny] at (\x, -0.12) {\label};
}

\draw[bar_corr] (0, \yBP+\gap) rectangle (7.3, \yBP+\gap+\bh);
\draw[bar_valid] (0, \yBP-\bh) rectangle (0.48, \yBP);
\node[anchor=east, font=\scriptsize] at (-0.15, \yBP) {BP: $\text{economy}_{L1}$};
\node[anchor=west, font=\tiny] at (7.4, \yBP+\gap+\bh/2) {0.73};
\node[anchor=west, font=\tiny] at (0.55, \yBP-\bh/2) {0.048};

\draw[bar_corr] (0, \yShell+\gap) rectangle (6.8, \yShell+\gap+\bh);
\draw[bar_valid] (0, \yShell-\bh) rectangle (0.47, \yShell);
\node[anchor=east, font=\scriptsize] at (-0.15, \yShell) {Shell: $\text{economy}_{L1}$};
\node[anchor=west, font=\tiny] at (6.9, \yShell+\gap+\bh/2) {0.68};
\node[anchor=west, font=\tiny] at (0.55, \yShell-\bh/2) {0.047};

\draw[bar_corr] (0, \yNexteraM+\gap) rectangle (5.2, \yNexteraM+\gap+\bh);
\draw[bar_valid] (0, \yNexteraM-\bh) rectangle (0.36, \yNexteraM);
\node[anchor=east, font=\scriptsize] at (-0.15, \yNexteraM) {NextEra: $\text{market}_{L2}$};
\node[anchor=west, font=\tiny] at (5.3, \yNexteraM+\gap+\bh/2) {0.52};
\node[anchor=west, font=\tiny] at (0.45, \yNexteraM-\bh/2) {0.036};

\draw[bar_corr] (0, \yNexteraI+\gap) rectangle (7.3, \yNexteraI+\gap+\bh);
\draw[bar_valid] (0, \yNexteraI-\bh) rectangle (0.35, \yNexteraI);
\node[anchor=east, font=\scriptsize] at (-0.15, \yNexteraI) {NextEra: $\text{inflation}_{L3}$};
\node[anchor=west, font=\tiny] at (7.4, \yNexteraI+\gap+\bh/2) {0.73};
\node[anchor=west, font=\tiny] at (0.45, \yNexteraI-\bh/2) {0.035};

\draw[bar_corr] (0, \yClearway+\gap) rectangle (4.5, \yClearway+\gap+\bh);
\draw[bar_valid] (0, \yClearway-\bh) rectangle (0.34, \yClearway);
\node[anchor=east, font=\scriptsize] at (-0.15, \yClearway) {Clearway: $\text{investors}_{L2}$};
\node[anchor=west, font=\tiny] at (4.6, \yClearway+\gap+\bh/2) {0.45};
\node[anchor=west, font=\tiny] at (0.45, \yClearway-\bh/2) {0.034};

\draw[bar_corr] (-5.6, -1.45) rectangle (-4.8, -1.15);
\node[anchor=west, font=\tiny] at (-4.7, -1.3) {Correlation $|r|$};

\draw[bar_valid] (-5.6, -2.15) rectangle (-4.8, -1.85);
\node[anchor=west, font=\tiny] at (-4.7, -2) {Validated $|\hat{\beta}|$};

\end{tikzpicture}
\caption{Comparison of correlational effect sizes (red, Pearson $|r|$) versus refutation-validated regression coefficients (green, $|\hat{\beta}|$ in daily return units). Raw correlations range from 0.45 to 0.73, while validated effects are an order of magnitude smaller (0.034--0.048), illustrating the substantial ``deflation'' that occurs when spurious associations are filtered through systematic robustness testing.}
\label{fig:effect-comparison}
\end{figure}

\subsection{Explainability Through Robustness-Validated Structure}

The superiority of our causal inference approach extends beyond statistical rigor to enhanced explainability. Each identified relationship carries an interpretable causal narrative grounded in economic theory. For instance, the positive effect of \emph{economy} sentiment on traditional energy stocks aligns with their role as cyclical assets whose valuations depend on economic growth expectations ~\citep{ziolo2024role, Kilian2020OilShocksHousing}. The coefficient magnitude (\(\approx 0.5\) bps) represents an economically meaningful daily impact that compounds to approximately 12 basis points monthly for sustained sentiment shifts.

In contrast, correlation-based metrics like uncertainty coefficients provide limited interpretability. While FinXABSA reports uncertainty coefficients up to 0.29, these information-theoretic measures lack the directional clarity and economic meaning of causal effects. A practitioner cannot determine from an uncertainty coefficient whether positive sentiment increases or decreases returns, nor can they quantify the economic magnitude of the relationship.

Our refutation framework also provides explicit confidence in causal claims. When we report that BP's response to economy sentiment passes all four refutation tests, this conveys specific guarantees: the effect is not due to multiple testing (placebo test), omitted variables (random common cause), outliers (subset stability), or distributional assumptions (bootstrap). This transparency enables practitioners to make informed decisions about which signals warrant trading strategies versus further investigation.

The temporal structure of effects offers additional insights. The concentration of significant effects at lags 1--2, with decay thereafter, suggests sentiment information is rapidly but not instantaneously incorporated into prices. This finding has practical implications for trading strategy design, indicating a narrow window for sentiment-based alpha generation that closes within 48 hours of information release.

\section{Discussions and Conclusion}
\label{sec:discussions_and_conclusion}

\subsection{Study Limitations}

The findings should be interpreted with caution due to a combination of contextual, methodological, and data-related constraints. The analysis is confined to a specific macroeconomic regime (Q4 2022), characterised by monetary tightening, energy-market disruptions, and heightened geopolitical uncertainty, which may limit temporal generalisability. Sentiment signals are derived exclusively from $\mathbb{X}$, whose demographic composition and subsequent platform-level changes may introduce selection bias and impede reproducibility. Finally, while extensive refutation tests were conducted, the observational nature of the study precludes full causal identification; unobserved confounders such as private information flows, algorithmic trading activity, and institutional rebalancing may partially account for the observed associations.

\subsection{From Correlation to Robust Association: What This Framework Does and Doesn't Claim}

We emphasize critical distinctions between our refutation-testing approach and genuine causal inference. 

\paragraph{What Our Framework Achieves:}
\begin{itemize}
\item \textbf{Filters spurious correlations} through systematic robustness checks (placebo tests, synthetic confounders, stability analysis)
\item \textbf{Establishes temporal precedence} by examining lagged sentiment predicting future returns
\item \textbf{Provides bounded confidence} in associations through multiple independent validation layers
\item \textbf{Yields economically interpretable} effect sizes with directional clarity
\end{itemize}

\paragraph{What Constitutes True Causal Inference:}
Establishing definitive causality requires addressing three fundamental challenges ~\citep{Pearl2009Causality}:
\begin{enumerate}
\item \textbf{Confounding:} Unobserved variables correlated with both sentiment and returns
\item \textbf{Reverse causality:} Returns potentially influencing subsequent sentiment expression
\item \textbf{Selection bias:} Non-random patterns in who tweets and when
\end{enumerate}

The random common cause test addresses confounding concerns by testing robustness to synthetic confounders, but cannot eliminate all unobserved variable bias. Temporal precedence mitigates but does not eliminate reverse causality concerns.

\paragraph{Appropriate Interpretation:} We identify \emph{refutation-validated predictive associations} that (1) survive multiple robustness checks, (2) exhibit temporal precedence, (3) align with economic theory, and (4) demonstrate effect sizes inconsistent with pure noise. These properties make identified signals suitable for trading strategies and risk management while acknowledging that precise causal mechanisms remain partially uncertain. Future work employing instrumental variables—such as exogenous sentiment shocks from natural disasters or regulatory announcements—could strengthen causal claims.

\subsection{Sample Size and Coverage Limitations}

\textbf{Our analysis is substantially constrained by sample size.} With only six stocks over a single quarter (Q4 2022), statistical power is limited and generalizability is uncertain. This sample size is insufficient for:
\begin{itemize}
\item Robust sector-wide conclusions about energy markets
\item Detection of heterogeneous treatment effects across firm characteristics
\item Panel methods with firm fixed effects that control for time-invariant confounders
\item Subgroup analysis comparing high-liquidity vs. low-liquidity stocks
\end{itemize}

Additionally, Q4 2022 represents a specific macroeconomic regime characterized by Federal Reserve tightening, energy price volatility, and post-pandemic market dynamics. Identified associations may be regime-dependent, with different patterns emerging during:
\begin{itemize}
\item Economic expansions vs. recessions
\item High vs. low volatility periods  
\item Different monetary policy stances
\item Energy supply shocks vs. stable periods
\end{itemize}

\textbf{We explicitly frame this study as an exploratory methodological proof-of-concept} demonstrating refutation-testing principles for ABSA-based financial analysis. Validation requires:
\begin{enumerate}
\item \textbf{Expanded cross-sectional coverage:} 15--30 stocks per sector minimum
\item \textbf{Extended time series:} Multi-year rolling-window estimation
\item \textbf{Out-of-sample validation:} Hold-out periods for predictive performance assessment
\item \textbf{Cross-sector replication:} Technology, healthcare, financial sectors
\end{enumerate}

Future work should implement these extensions before drawing policy or investment recommendations.

\subsection{Methodological Constraints}

Despite comprehensive refutation testing, several methodological limitations warrant consideration. First, our sentiment measurement relies on aspect-level aggregation that may obscure within-day dynamics and intraday sentiment-return relationships. High-frequency analysis using tick data could reveal microstructure effects invisible at daily frequencies. Second, the linear specification may miss important non-linearities and threshold effects. Sentiment impact might exhibit asymmetry between positive and negative domains or state-dependence conditional on market volatility ~\citep{Garcia2013SentimentRecessions}.

The HAC standard errors, while addressing serial correlation, assume stationarity that may be violated during crisis periods. Future research should explore time-varying parameter models and regime-switching frameworks that allow causal effects to evolve with market conditions. Additionally, our univariate approach examines each aspect independently, potentially missing interaction effects where multiple aspects jointly influence returns.

\subsection{Implications for Regulatory Compliance and Explainable AI}

The European Union's AI Act, which entered into force in August 2024, establishes stringent transparency and explainability requirements for high-risk AI systems in financial services ~\citep{kim2025ai}. Credit scoring, risk assessment, and algorithmic trading systems face particular scrutiny, with regulators requiring that automated decisions be interpretable, auditable, and free from unjustified bias ~\citep{habibullah2024explainable}.
Our refutation-testing framework directly addresses these regulatory demands. Unlike black-box sentiment aggregators that produce opaque signals, our approach provides:

\begin{enumerate}
    \item \textbf{Transparent assumptions:} Each association is explicitly conditional on specified confounders and lag structures
    \item \textbf{Auditable validation:} Refutation test results provide documented evidence that signals survive multiple robustness checks
    \item \textbf{Directional interpretability:} Effect sizes carry clear economic meaning (basis points per standard deviation) rather than abstract correlation coefficients
    \item \textbf{Bounded confidence:} Bootstrap intervals and subset stability tests quantify uncertainty in ways amenable to risk management frameworks
\end{enumerate}

The Financial Stability Oversight Council's 2024 Annual Report elevated AI as a systemic risk concern, specifically citing model opacity and the difficulty of auditing complex ML systems ~\citep{FSOC2024Report}. Refutation testing offers a practical middle ground between fully interpretable linear models (which may sacrifice predictive power) and opaque deep learning systems (which face regulatory skepticism). By demonstrating that sentiment signals can be rigorously validated without sacrificing economic meaning, our framework provides a template for compliance-ready sentiment analytics.

For practitioners deploying sentiment-based strategies, we recommend:
\begin{itemize}
    \item Documenting refutation test results as part of model validation packages
    \item Establishing minimum pass rates across refutation suites before signal deployment
    \item Maintaining audit trails linking trading decisions to specific validated associations
    \item Implementing periodic re-validation to detect regime changes that may invalidate historical relationships
\end{itemize}

\subsection{Causal Identification Challenges}

While refutation tests strengthen causal claims, they cannot eliminate all threats to identification. Unobserved confounders correlated with both sentiment and returns—such as private information diffusion or algorithmic trading patterns—may bias estimates despite passing refutation tests. The assumption of no anticipatory effects (strict exogeneity) may be violated if sophisticated traders act on sentiment predictions before they materialise in social media.

Furthermore, our framework assumes homogeneous treatment effects across stocks within each sector. Heterogeneous responses based on firm characteristics (size, leverage, analyst coverage) could provide more granular insights but require larger samples for reliable estimation. Panel methods with firm fixed effects and clustered standard errors represent a natural extension.

\subsection{Limitations}

The findings presented in this study are subject to several important constraints. The small sample (six stocks, one quarter) precludes definitive sector-wide conclusions. Our refutation tests strengthen confidence in identified associations but cannot establish causality without instrumental variables or natural experiments. Unobserved confounders---such as private information flows, algorithmic trading activity, and institutional rebalancing---may partially account for the observed relationships, and the observational nature of the study precludes full causal identification. We therefore frame contributions as methodological, demonstrating how systematic robustness testing filters spurious correlations in high-dimensional sentiment analysis, rather than as definitive empirical claims about energy markets.

\subsection{Future Research Directions}

Several avenues merit investigation. First, incorporating large language models (LLMs) for sentiment extraction could improve aspect detection and sentiment classification accuracy. Zero-shot and few-shot learning approaches might identify emergent aspects not captured by predefined lexicons. Second, graph neural networks could model the interdependence structure between aspects, capturing sentiment spillovers and contagion effects.

Methodologically, instrumental variable approaches using exogenous sentiment shocks (regulatory announcements, natural disasters) could provide stronger identification. Synthetic control methods ~\citep{bouttell2018synthetic, Abadie2010SyntheticControl} comparing treated stocks to synthetic counterfactuals offer another identification strategy. Machine learning methods for causal inference, including causal forests ~\citep{Wager2018CausalForests} and double machine learning ~\citep{Chernozhukov2018DML}, could accommodate high-dimensional controls while maintaining valid inference.

From a practical perspective, developing real-time implementation requires addressing several challenges: sentiment extraction latency, online learning for parameter updates, and transaction cost modelling. Integration with portfolio optimisation frameworks would translate causal insights into implementable trading strategies with explicit risk-return tradeoffs.

Extending the framework to other textual sources---earnings calls, analyst reports, regulatory filings---would provide a more comprehensive view of information flow in financial markets. Cross-modal analysis combining text, audio, and visual data represents the frontier for multimodal financial sentiment analysis, requiring new causal frameworks for heterogeneous data integration.

\subsection{Conclusion}

This paper presented a \textbf{robustness-testing framework} that elevates aspect-based sentiment analysis beyond naive correlation by combining aspect-specific signals (net-ratio scoring with z-normalisation), OLS with Newey--West HAC errors, and four refutation tests (placebo, random common cause, subset stability, bootstrap) to filter spurious associations. Applied to \(\sim\)120{,}000 $\mathbb{X}$ posts in Q4 2022 for six energy-sector equities, the framework isolates a small set of \textbf{refutation-validated associations} that are economically interpretable and statistically robust: \emph{economy} sentiment predicts next-day returns for BP and Shell (\(\approx\)0.48/0.47 bps per s.d.), while renewables exhibit aspect- and horizon-specific responses. The framework yields sentiment signals that are directionally interpretable, economically sized, and more reliable than correlational baselines, providing a foundation for expanded validation across markets and time periods.

\section*{Acknowledgements}
This research is supported by the RIE2025 Industry Alignment Fund -- Industry Collaboration Projects (IAF-ICP) (Award I2301E0026), administered by A*STAR, as well as supported by Alibaba Group and NTU Singapore through Alibaba-NTU Global e-Sustainability CorpLab (ANGEL). The work is also supported by the Ministry of Education, Singapore under its MOE Academic Research Fund Tier 2 (MOE-T2EP20123-0005).

\appendix
\renewcommand{\thefigure}{A\arabic{figure}}
\renewcommand{\thetable}{A\arabic{table}}
\section{Appendix}
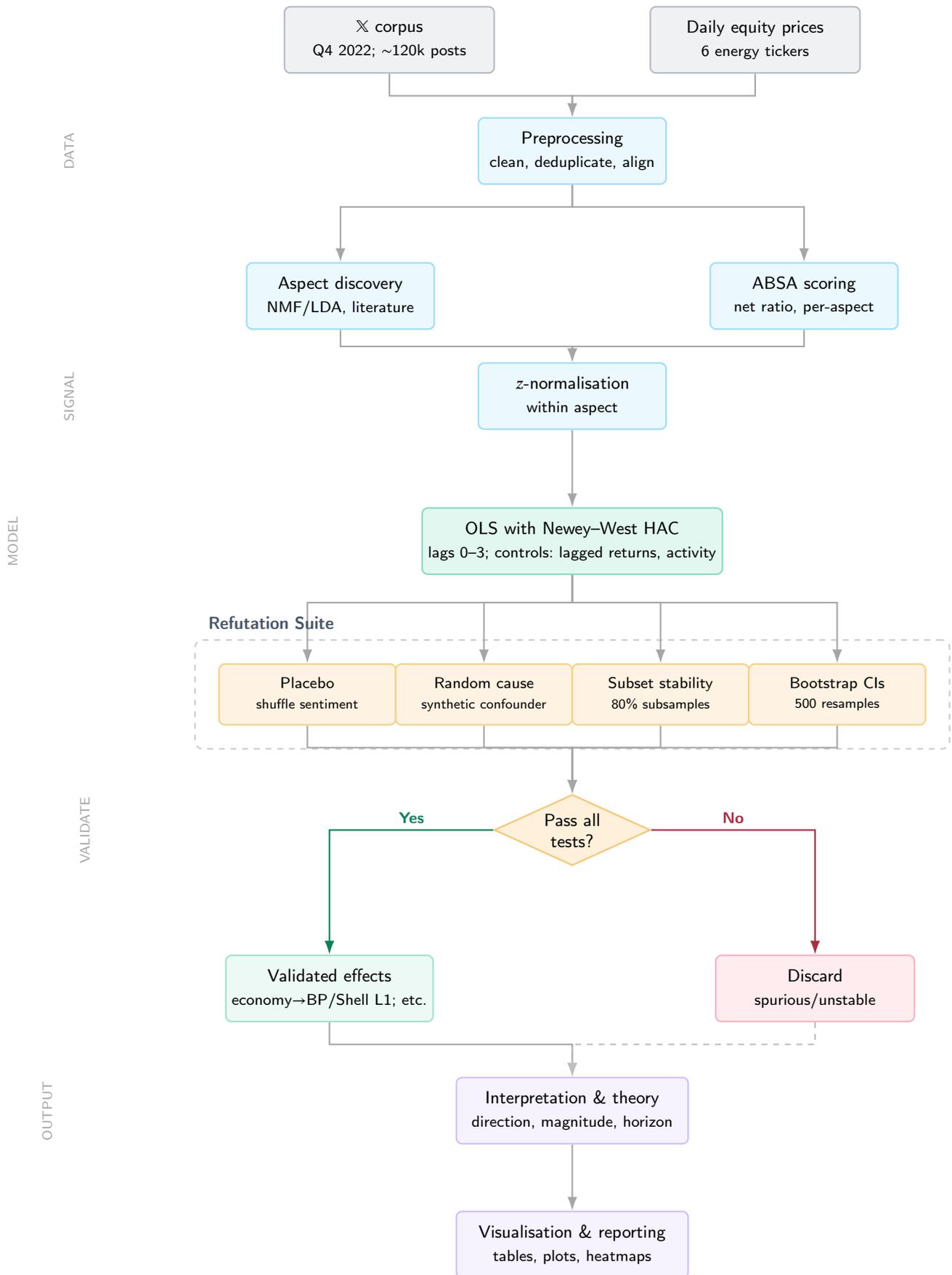
\begin{figure*}[t]
  \centering
  \begin{tikzpicture}[
    node distance=10mm and 14mm,
    basebox/.style={
      rectangle, 
      rounded corners=3pt, 
      draw=gray!60, 
      thick, 
      align=center, 
      font=\small,
      minimum height=12mm,
      text depth=0.5ex
    },
    databox/.style={
      basebox,
      minimum width=38mm,
      fill=slate!8,
      draw=slate!40
    },
    processbox/.style={
      basebox,
      minimum width=34mm,
      fill=sky!10,
      draw=sky!50
    },
    modelbox/.style={
      basebox,
      minimum width=50mm,
      fill=emerald!12,
      draw=emerald!50
    },
    refutebox/.style={
      basebox,
      minimum width=32mm,
      minimum height=11mm,
      fill=amber!12,
      draw=amber!50,
      font=\footnotesize
    },
    outcomebox/.style={
      basebox,
      minimum width=36mm,
      fill=emerald!8,
      draw=emerald!40
    },
    rejectbox/.style={
      basebox,
      minimum width=36mm,
      fill=rose!10,
      draw=rose!40
    },
    interpretbox/.style={
      basebox,
      minimum width=42mm,
      fill=violet!8,
      draw=violet!40
    },
    decision/.style={
      diamond, 
      aspect=2.2, 
      draw=amber!60, 
      thick, 
      align=center, 
      fill=amber!15,
      font=\small,
      inner sep=2pt
    },
    groupbox/.style={
      rectangle, 
      rounded corners=4pt, 
      draw=gray!40, 
      thick,
      dashed,
      inner sep=10pt
    },
    arrow/.style={
      -{Latex[length=2.5mm, width=1.8mm]}, 
      thick, 
      color=gray!70
    },
    yespath/.style={
      arrow,
      color=emerald!70!black
    },
    nopath/.style={
      arrow,
      color=rose!70!black
    },
    label/.style={
      font=\scriptsize,
      color=gray!70
    }
  ]
  
  \definecolor{slate}{RGB}{71,85,105}
  \definecolor{sky}{RGB}{56,189,248}
  \definecolor{emerald}{RGB}{16,185,129}
  \definecolor{amber}{RGB}{245,158,11}
  \definecolor{rose}{RGB}{244,63,94}
  \definecolor{violet}{RGB}{139,92,246}

  \node[databox] (tweets) {$\mathbb{X}$ corpus\\[1pt]{\footnotesize Q4 2022; $\sim$120k posts}};
  \node[databox, right=28mm of tweets] (prices) {Daily equity prices\\[1pt]{\footnotesize 6 energy tickers}};
  
  \node[processbox, below=14mm of $(tweets)!0.5!(prices)$] (prep) {Preprocessing\\[1pt]{\footnotesize clean, deduplicate, align}};

  \node[processbox, below=14mm of prep, xshift=-42mm] (aspects) {Aspect discovery\\[1pt]{\footnotesize NMF/LDA, literature}};
  \node[processbox, below=14mm of prep, xshift=42mm] (sent) {ABSA scoring\\[1pt]{\footnotesize net ratio, per-aspect}};
  \node[processbox, below=12mm of $(aspects)!0.5!(sent)$] (zscore) {$z$-normalisation\\[1pt]{\footnotesize within aspect}};

  \node[modelbox, below=14mm of zscore] (model) {OLS with Newey--West HAC\\[1pt]{\footnotesize lags 0--3; controls: lagged returns, activity}};

  \node[refutebox, below=16mm of model, xshift=-48mm] (placebo) {Placebo\\[1pt]{\scriptsize shuffle sentiment}};
  \node[refutebox, below=16mm of model, xshift=-16mm] (rcc) {Random cause\\[1pt]{\scriptsize synthetic confounder}};
  \node[refutebox, below=16mm of model, xshift=16mm] (subset) {Subset stability\\[1pt]{\scriptsize 80\% subsamples}};
  \node[refutebox, below=16mm of model, xshift=48mm] (boot) {Bootstrap CIs\\[1pt]{\scriptsize 500 resamples}};
  
  \begin{pgfonlayer}{bg}
    \node[groupbox, fit=(placebo)(rcc)(subset)(boot), inner sep=12pt] (refgroup) {};
  \end{pgfonlayer}
  \node[font=\small\bfseries, color=slate, anchor=south west] at ([xshift=4pt, yshift=2pt]refgroup.north west) {Refutation Suite};

  \node[decision, below=18mm of $(rcc)!0.5!(subset)$] (pass) {Pass all\\tests?};

  \node[outcomebox, below=16mm of pass, xshift=-44mm] (findings) {Validated effects\\[1pt]{\footnotesize economy$\to$BP/Shell L1; etc.}};
  \node[rejectbox, below=16mm of pass, xshift=44mm] (reject) {Discard\\[1pt]{\footnotesize spurious/unstable}};

  \node[interpretbox, below=16mm of $(findings)!0.5!(reject)$] (explain) {Interpretation \& theory\\[1pt]{\footnotesize direction, magnitude, horizon}};
  \node[interpretbox, below=12mm of explain] (report) {Visualisation \& reporting\\[1pt]{\footnotesize tables, plots, heatmaps}};

  \begin{pgfonlayer}{bg}
    \draw[arrow] (tweets.south) -- ++(0,-4mm) -| (prep.north);
    \draw[arrow] (prices.south) -- ++(0,-4mm) -| (prep.north);
    
    \draw[arrow] (prep.south) -- ++(0,-4mm) -| (aspects.north);
    \draw[arrow] (prep.south) -- ++(0,-4mm) -| (sent.north);
    
    \draw[arrow] (aspects.south) -- ++(0,-3mm) -| (zscore.north);
    \draw[arrow] (sent.south) -- ++(0,-3mm) -| (zscore.north);
    
    \draw[arrow] (zscore) -- (model);
    
    \draw[arrow] (model.south) -- ++(0,-5mm) -| (placebo.north);
    \draw[arrow] (model.south) -- ++(0,-5mm) -| (rcc.north);
    \draw[arrow] (model.south) -- ++(0,-5mm) -| (subset.north);
    \draw[arrow] (model.south) -- ++(0,-5mm) -| (boot.north);
    
    \draw[arrow] (placebo.south) -- ++(0,-4mm) -| (pass.north);
    \draw[arrow] (rcc.south) -- ++(0,-4mm) -| (pass.north);
    \draw[arrow] (subset.south) -- ++(0,-4mm) -| (pass.north);
    \draw[arrow] (boot.south) -- ++(0,-4mm) -| (pass.north);
    
    \draw[yespath] (pass.west) -| node[above, pos=0.25, font=\footnotesize\bfseries, color=emerald!70!black] {Yes} (findings.north);
    \draw[nopath] (pass.east) -| node[above, pos=0.25, font=\footnotesize\bfseries, color=rose!70!black] {No} (reject.north);
    
    \draw[arrow] (findings.south) -- ++(0,-4mm) -| (explain.north);
    \draw[arrow, dashed, gray!50] (reject.south) -- ++(0,-4mm) -| (explain.north);
    
    \draw[arrow] (explain) -- (report);
  \end{pgfonlayer}

  \node[label, rotate=90, anchor=south] at ([xshift=-72mm]prep.west) {DATA};
  \node[label, rotate=90, anchor=south] at ([xshift=-72mm]zscore.west) {SIGNAL};
  \node[label, rotate=90, anchor=south] at ([xshift=-72mm]model.west) {MODEL};
  \node[label, rotate=90, anchor=south] at ([xshift=-72mm]pass.west) {VALIDATE};
  \node[label, rotate=90, anchor=south] at ([xshift=-72mm]explain.west) {OUTPUT};

  \end{tikzpicture}
  \caption{End-to-end workflow for refutation-validated sentiment analysis. Data inputs (top) flow through signal construction and OLS modelling to a four-part refutation suite. Only associations passing \emph{all} refutation tests proceed to interpretation; spurious signals are discarded.}
  \label{fig:workflow}
\end{figure*}

Please see Figure \ref{fig:workflow}.

\printcredits

\bibliographystyle{cas-model2-names}

\bibliography{references}

\end{document}